\documentclass[12pt]{article}

\usepackage[utf8x]{inputenc}
\input{glyphtounicode}
\usepackage{amsmath}
\usepackage{graphicx}
\usepackage{xspace}
\usepackage[font=small,labelfont=bf]{caption}
\usepackage{subcaption}
\usepackage{setspace}
\usepackage{color}
\usepackage{soul}
\usepackage[section]{placeins}
\usepackage{enumitem}
\usepackage{float}
\setlist[enumerate]{itemsep=-1mm}
\setlist[itemize]{itemsep=-1mm}
\usepackage[figure, table]{totalcount}
\usepackage{lastpage}
\usepackage{lineno}
\usepackage{marginnote}
\usepackage{amsfonts}
\usepackage{algorithm}
\usepackage[]{algorithmicx, setspace}
\usepackage{algpseudocode}
\usepackage[numbers]{natbib}

\DeclareRobustCommand{\figref}[1]{{\textcolor{black}{Fig \ref{#1}}}}
\DeclareRobustCommand{\eqref}[1]{{\textcolor{black}{Eq \ref{#1}}}}

\author{Kristjan Kalm \\
\\
\normalsize{MRC Cognition and Brain Sciences Unit, University of Cambridge}\\
\normalsize{15 Chaucer Road, Cambridge, CB2 7EF, UK}\\
\\
\normalsize{E-mail: kristjan.kalm@mrc-cbu.cam.ac.uk.}
}

\begin{document}
\title{Recency-weighted Markovian inference}
\maketitle
\onehalfspacing

\subsection*{Abstract}

We describe a Markov latent state space (MLSS) model, where the latent state distribution is a decaying mixture over multiple past states. We present a simple sampling algorithm that allows to approximate such high-order MLSS with fixed time and memory costs.

\section{Introduction} 
\label{sec:intro}

Markovian inference methods allow for on-line processing of data with time and memory requirements that are constant in the total number of observations received at each time step. Each observation $y_t$ corresponds to a latent variable $z_t$, which over time forms a Markov chain, giving rise to the latent state space model \cite{Bishop2006b}. At every time step $t$ a new model $z_t$ is inferred using the Bayes theorem based on the previous state of the model $z_{t-1}$ and current observation $y_t$:
\begin{equation}
\label{eq:bayes}
	p(z_t|y_t) \propto p(y_t|z_t) \cdot p(z_t|z_{t-1}),
\end{equation}

where the probability distribution (or density) $p(z_t|z_{t-1})$ describes a \emph{state transition function} responsible for the evolution of the latent state $z$. However, the fixed computational cost of such inference depends on assuming conditional independence between current and past $m$ states of the latent variable so that:
\begin{align}
	\label{eq:cond_ind}
	p(z_t|z_{t-1}, \dots, z_{t-m}) = p(z_t|z_{t-1}).
\end{align}

Contrastingly, in many human learning phenomena the model of the environment seems to be inferred across multiple past states and weighed with respect to their recency (see \cite{Kiyonaga2017} for a review). If we want to weigh the effect of each past $m$ states separately on the current state $z_t$ then the Markov property (\eqref{eq:cond_ind}) does not hold and we need to find another way of defining the evolution of the latent state. To solve this problem we use the \emph{mixture state transition function} approach for high-order Markov chains \cite{Berchtold2002,Saul1999}. This approach represents the conditional probability distribution $p(z_t|z_{t-1}, \dots, z_{t-m})$ as a mixture of past $m$ states:
\begin{equation}
	\label{eq:mix1}
	p(z_t|z_{t-1}, \dots, z_{t-m}) = \theta_1 p(z_t|z_{t-1}) + \dots + \theta_m p(z_t|z_{t-m}) = \sum^{M}_{m=1} \theta_m p(z_t|z_{t-m}) ,
\end{equation}

where $\theta$ is a mixing coefficient so that 
\[ 0 < \theta_m \leq 1 \text{, and } \sum^{M}_{m=1} \theta_m = 1. \]

Mixing coefficients make the dependence of the future on the past explicit by quantifying the decay in dependence as the future moves farther from the past \cite{McDonald2011}. If we use an equal mixing coefficient for all previous states, $\theta=1/m$, then all past states, independent of their lag, contribute equally to the prediction of the current state. Instead, we might want the more recent states represented proportional to their recency. Hence, we can assume that the contribution of past states, reflected by the mixing coefficient $\theta$, declines over $m$ time steps as given by some decay function $\phi$ and the rate of decay parameter $\beta$. Here we choose a decay function $\phi$, so that:
\begin{equation}
	\label{eq:decay}
	\theta_m = \phi(m, \beta) = \alpha \theta_{0} (1-\beta)^{m}, 
\end{equation}

where $\beta$ is the rate of decrease ($ 0 \leq \beta < 1$) and $\alpha$ normalising constant. Substituting $\theta_m$ into equation (\eqref{eq:mix1}) gives:
\begin{equation}
	\label{eq:mix}
	p(z_t|z_{t-1}, \dots, z_{t-m}) = \sum^{M}_{m=1} \alpha \theta_{0} (1-\beta)^{m} p(z_t|z_{t-m}).
\end{equation}

\begin{figure}[!ht]
\centering
	\caption{Values of the mixing coefficient $\theta$ over past 5 states ($z_{t-1}, \dots , z_{t-5}$) based on different  $\beta$ values. $\theta_t$ represents the proportion of a past state $z_t$ in the mixture distribution (\eqref{eq:mix}).}
	\includegraphics[width=0.5\textwidth]{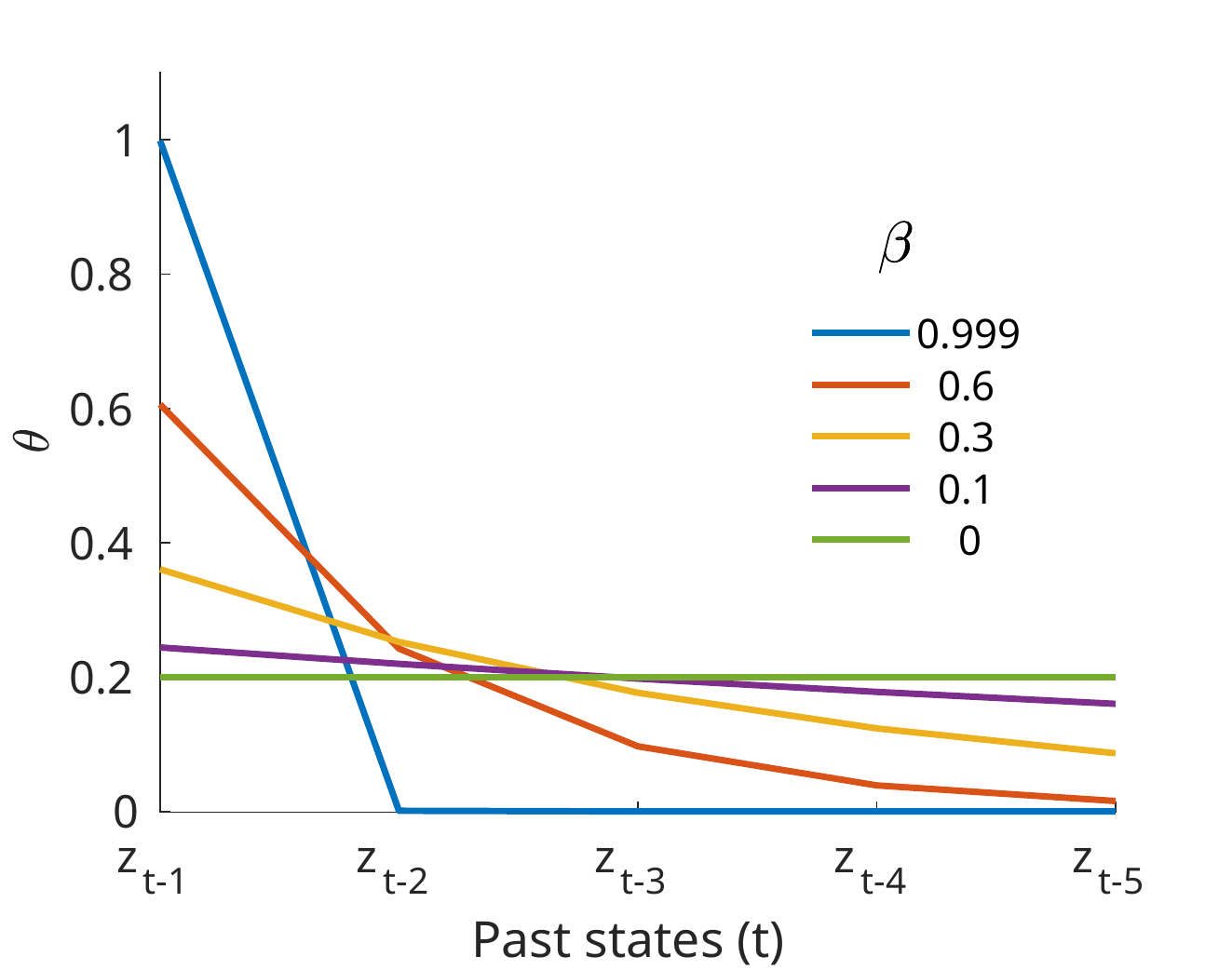}	
	\label{fig:beta}
\end{figure}

As a result we have a decaying time window into the past $m$ states defined by the rate parameter $\beta$ of the decreasing mixing coefficient $\theta$. \figref{fig:beta} illustrates the relationship between the $\beta$ and $\theta$ parameters: the bigger the $\beta$ the faster the utility of past states decreases and greater the contributions of most recent states to the mixture distribution (\eqref{eq:mix}). As $\beta$ approaches 1, the mixture begins to resemble $p(z_t|z_{t-1})$ and approximate a first-order Markov chain:
\[ \lim_{\beta \to 1}, p(z_t|z_{t-1}, \dots, z_{t-m}) = p(z_t|z_{t-1}). \]

Conversely, as $\beta$ approaches 0, the mixing coefficient $\theta$ does not decay across time steps and all past states contribute equally to the mixture. Intuitively, $\beta$ could be interpreted as the bias towards more recent states. 

However, such an approach poses two fundamental problems. First, to evaluate \eqref{eq:mix} we still need to explicitly store all past $m$ states and the processing time of the algorithm would increase with $m$. Second, such sequentially estimated mixtures are themselves mixtures of previous mixture distributions, making the estimation analytically intractable. Therefore, such a model would not allow for processing of data with time and memory requirements that are constant at each time step. 

Next, we show how a practical solution to both of these problems can be obtained by using sampling methods. 

\section{State estimation with mixture sampling}
In order to limit the computational cost of performing inference at every time step we represent the distribution of latent variable $z$ with a fixed number of samples $L$. As a result the proportion of samples assigned to a particular component of the mixture distribution (representing a past state) is determined by the mixing coefficient $\theta$:
\begin{equation}
	\label{eq:samp}
	p(z_t|z_{t-1}, \dots, z_{t-m}) = \sum^{M}_{m=1} \theta_m p(z_t|z_{t-m}) \simeq \sum^{M}_{m=1} \sum^{\theta_m L}_{l=1}\{p(z_t|z_{t-m})\}^{(l)}
\end{equation}

where $L$ is the total number of samples, $\theta_m L$ a subset of samples allocated to $p(z_t|z_{t-m})$, $\{p(z_t|z_{t-m})\}$ a set of samples from that distribution, and $\theta L_m$ is rounded to the nearest integer. 

For example, if we use $L=100$ samples to represent a mixture of $m=5$ components with a fixed $\theta=1/5$, then every component would be assigned 20 samples. With a decaying mixing coefficient (\eqref{eq:decay}) the number of samples assigned to any $z_{t-m}$ decreases with $m$, but the number of samples assigned to any particular $z_{t-m}$ will remain constant (because of the constant rate parameter $\beta$ in \eqref{eq:mix}; also see \figref{fig:beta}). For example, if $t>m$, $L=100$, $\beta=0.5$, and $\theta_0=1$, then $p(z_t|z_{t-1})$ will be represented with 50 samples, $p(z_t|z_{t-2})$ with 25 samples, and so forth.

The property of constant number of samples for every $m$-th component of the mixture at any time is important since it greatly simplifies the approximation of the mixture distribution (\eqref{eq:mix}). If at every time-step $t$ we choose a fixed proportion $\beta$ of samples from the existing mixture distribution and reassign those samples to represent $p(z_t|z_{t-1})$, then after $m$ steps we end up with samples allocated across mixture components as given by (\eqref{eq:mix}). To make this explicit, consider the following evolution of mixture (\eqref{eq:mix}) symbolically in a table:

\bgroup
\def\arraystretch{1.5}
\begin{table}[H]
	\caption{The evolution of the latent state $z$ via mixture state transition function.  The left column shows the evolution of the mixture representation, the right column shows the same as a sample representation. Sample representation: at every time-step $t$ we choose a fixed proportion $\beta$ of samples and reassign those samples to represent the most recent mixture component. Here $z_0$ is the initial distribution, \{$\mathbf{z}_t\}^{(L)}$ is a set of $L$ samples representing the mixture distribution at time $t$, and $\{p(z)\}^{(l)}$ denotes a set of $l$ samples from $p(z)$.}
	\begin{tabular}{l| l| l }

	$t$ 
	& Mixture representation (\eqref{eq:mix1}) 
	& Sample representation \\
	\hline 
	1 & $\theta_1 p(z_1|z_0) + \theta_2 p(z_0)$		
	& $\{\mathbf{z}_1\}^{(L)} = \{ p(z_1|z_0) \}^{(L \beta)} + \{z_0\}^{(L(1-\beta))}$ \\
	 
	2 & $\theta_1 p(z_2|z_1) + \theta_2 p(z_1|z_0) + \theta_3 p(z_0)$		
	& $\{\mathbf{z}_2\}^{(L)} = \{ p(z_2|z_1) \}^{(L \beta)} + \{\mathbf{z}_1\}^{(L(1-\beta))}$ \\
	 
	3 & $\theta_1 p(z_3|z_2) + \theta_2 p(z_2|z_1) + \theta_3 p(z_1|z_0) + \theta_4 p(z_0)$
	& $\{\mathbf{z}_3\}^{(L)} = \{ p(z_3|z_2) \}^{(L \beta)} + \{\mathbf{z}_2\}^{(L(1-\beta))}$ \\

	& \dots & \dots \\

	t & $\theta_1 p(z_t|z_{t-1}) + \theta_{2} p(z_{t-1}|z_{t-2}) + \dots + \theta_m p(z_0)$
	& $\{\mathbf{z}_t\}^{(L)} = \{p(z_t|z_{t-1})\}^{(L \beta)} + \{\mathbf{z}_{t-1}\}^{(L(1-\beta))} $ \\

	\end{tabular}
	\label{table:evol}

\end{table}

\egroup

It follows that at every time step $t$ a mixture distribution of past $m$ states can be approximated by sampling from just $p(z_t|z_{t-1})$ and the previous state of the mixture $\{\mathbf{z}_{t-1}\}$. 

\begin{algorithm}[ht!]
	\setstretch{1.25}
	\caption{State estimation with mixture sampling}
	\label{alg:z}
	\begin{algorithmic}
		\State $t=0$
		\State$\{\mathbf{z}_0\}^{(L)} \sim p(z)$
		\Comment{Initialise $L$ by sampling from some distribution $p(z)$.}
		\Repeat 
		\State{t = t+1} 
		\State $\{\mathbf{z}_t\}^{(L)} = \{p(z_t|z_{t-1})\}^{(L \beta)} + \{\mathbf{z}_{t-1}\}^{(L(1-\beta))} $ 
		\Comment{Update $\{\mathbf{z}_t\}$ by taking $L \beta$ samples from \\ \hspace{245pt} $p(z_t|z_{t-1})$ and $L(1-\beta)$ samples from $ \{\mathbf{z}_{t-1}\}$}
		\Until{t=T}
	\end{algorithmic}
\end{algorithm}

After $m$ steps these $L$ samples represent a mixture of past $m$ states. The number of samples assigned to a particular mixture component  decreases with its distance from the present state at the rate of $\beta$. This algorithm has constant time and memory requirements and uses only two parameters: re-sampling coefficient $\beta$ and the number of total samples $L$. 

\section{Posterior estimation}

So far we have just dealt with high-order chains of latent variables $z$. However, our goal is to incrementally infer the model of the environment $z_t$ every time a new observation $y_t$ arrives. In other words, we are interested in inferring the posterior probability of the latent variable:
\[ p(z_t|y_t) \propto p(y_t|z_t) \cdot p(z_t|z_{t-1}, \dots, z_{t-m}). \]

We assume that the evolution of the latent variable $z$ is predicted by the mixture state transition function $h()$ and system noise $q$:
\begin{equation}
	\label{eq:state_transition}
	z_t = h(z_{t-1}, \dots, z_{t-m}) + q_t
\end{equation}

Having previously derived an approximation for $p(z_t|z_{t-1}, \dots, z_{t-m})$ sampling from the posterior distribution becomes straightforward. Here we use a sequential Monte Carlo approach called \emph{particle filtering} \cite{Cappe2007,Doucet2001}, where the estimation of the the posterior distribution is based on generating proposals of the latent state $p(z_t|z_{t-1}, \dots, z_{t-m})$ with Algorithm \ref{alg:z} and weighing them with the likelihood function $p(y_t|z_t)$. The process can be summarised as follows. 

At every time-step $t$ a set of $L$ particles $\{\mathbf{z}_{t}\}^{(L)}$ represents the mixture of past states (\eqref{eq:samp}). In other words, each individual particle $z_t^l$ is a prediction of the latent state at the current time step. Next we calculate the likelihood of the current observation $y_t$ under each particle. This likelihood serves as the importance weight $w_t^l$. We then normalise the importance weights across particles. The set of particles $\{\mathbf{z}_{t}\}^{(L)}$ together with the corresponding weights $\{\mathbf{w}_{t}\}^{(L)}$ represents the posterior distribution $p(z_t|y_t)$. Next, we create a new predictive distribution for the next state $p(z_{t+1}|z_t)$ represented by $\{\mathbf{z}_{t+1}\}^{(L)}$ by replacing randomly $L \beta$ particles in $\{\mathbf{z}_{t}\}^{(L)}$ with samples from the posterior $p(z_t|y_t)$. Finally we add system noise $q_t$ to the particles. This process is described fully in Algorithm \ref{alg:zy} below.

\begin{figure}[!ht]
\centering
	\caption{Components of the state mixture distribution for different values of the $\beta$ parameter (horisontal axis). When $\beta = 1$ the predictive prior for the next state consists entirely of samples from the last state $z_{t-1}$. As $\beta$ approaches zero the mixture distribution will contain roughly equal amount of samples from all of the past states.}
	\includegraphics[width=0.6\textwidth]{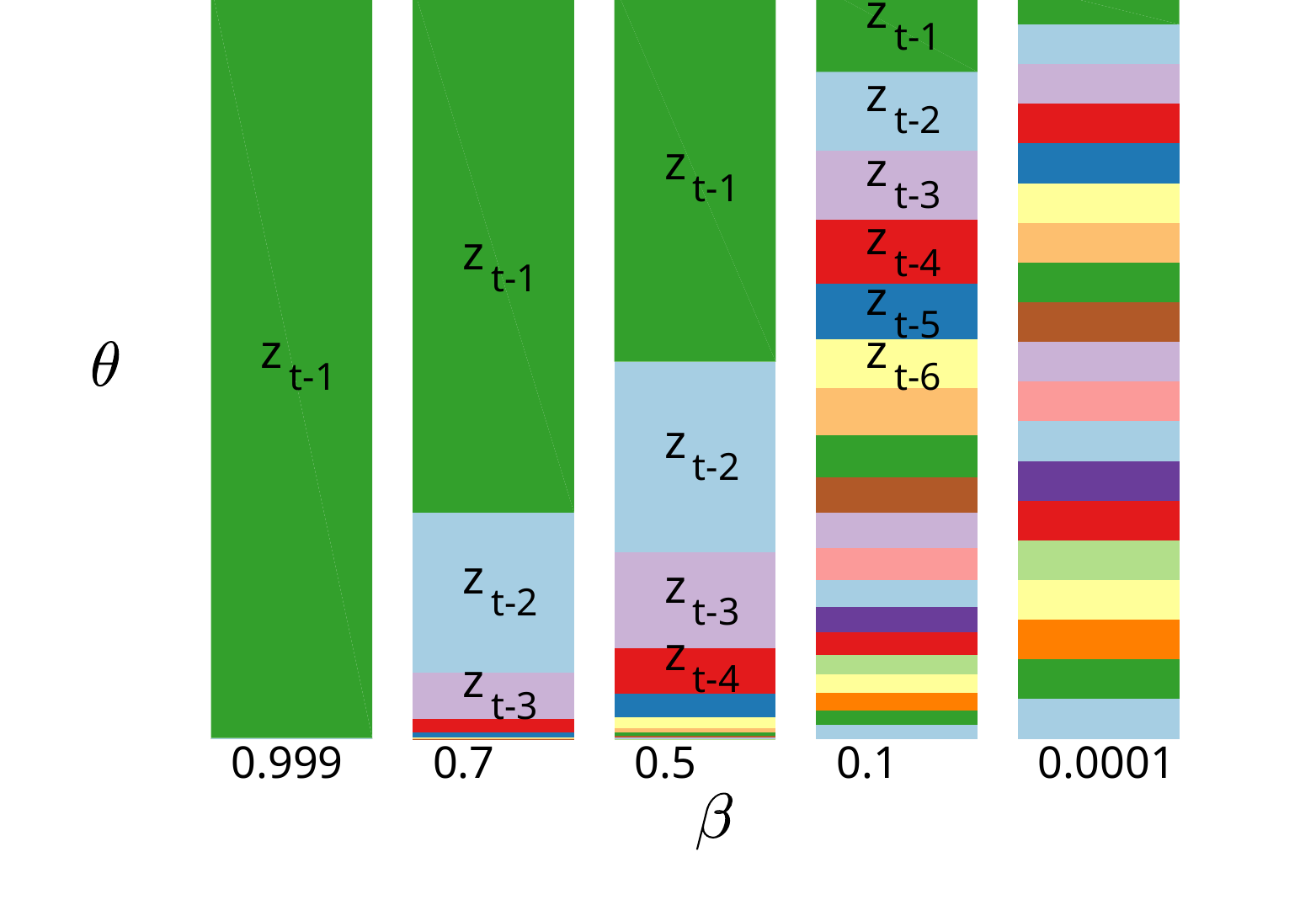}	
	\label{fig:theta}
\end{figure}

This algorithm is based on a widely used method of importance sampling: if we change the re-sampling step so that the predictive prior $\{\mathbf{z}_{t+1}\}^{(L)}$ is re-sampled solely from the posterior distribution $(z_t|y_t)$ by setting $\beta = 1$, then the algorithm becomes equivalent to the standard importance resampling implementation of a 1st order Markov chain of latent states \cite{Rubin1987}, where the system transition function is identity and system noise is zero (in other words, the posterior distribution of the current state $p(z_t|y_t)$ serves as the predictive prior for the next state $p(z_{t+1}|z_t)$, \figref{fig:theta}, column 1). 
\begin{algorithm}[ht!]
	\setstretch{1.7}
	\caption{Importance sampling}
	\label{alg:zy}
	\begin{algorithmic}
		\State $t=0$
		\State $\{\mathbf{z}_0\}^{(L)} \sim f(z)$ 
		\Comment{Initialise $L$ particles by sampling from some distribution $f(z)$.}
		\State $\{\mathbf{z}_1\}^{(L)} = \{\mathbf{z}_0\}^{(L)}$
		\Comment{Use $z_0$ as a prediction for $z_1$}
		\Repeat 
		\State{t = t+1} 
		\State $\{\mathbf{w}_t\}^{(L)} = p(y_t|\{\mathbf{z}_t\}^{(L)})$
		\Comment{Assign weights to samples}		
		\State $\{\mathbf{w}_t\}^{(L)} = \{\mathbf{w}_t\}^{(L)} / \sum \{{\mathbf{w}}_t\}^{(L)}$
		\Comment{Normalise weights}
		\State $ \{\mathbf{z}_t^{(i)}\}_{1 \leq i \leq N }^{(L \beta)}$ 
		\Comment{Randomly select $\beta L$ particle indices $j_i \in \{1, \dots, N\}$}
		\State $\{\mathbf{z}_t^{(i)}\}_{1 \leq i \leq N }^{(L \beta)} = \{p(z_t|y_t)\}^{(L\beta)} \quad$
		\Comment{Assign selected particles to samples from the posterior}		
		\State $\{\mathbf{z}_{t+1}\}^{(L)} = \{p(z_t|y_t)\}^{(L\beta)} + \{\mathbf{z}_t\}^{(L(1-\beta))} \quad $
		\Comment{New predictive latent distribution is a mixture of current posterior and previous mixture}
		\State $\{\mathbf{z}_{t+1}\}^{(L)} = \{\mathbf{z}_{t+1}\}^{(L)} + \mathbf{q}$ 
		\Comment{Add system noise to the particles}
		\Until{t=T}
	\end{algorithmic}
\end{algorithm}

\section{Conclusions}

In many human incremental learning situations the model of the environment is inferred across multiple past states and weighed with respect to their recency \cite{Kiyonaga2017}. Here we presented a probabilistic model of incremental learning where the current latent state given some observation $p(z_t|y_t)$ depends on recency-weighted past states. Despite the fact that our model approximates high-order dependencies we have described a posterior distribution estimation algorithm which is itself 1st order Markovian and can hence be processed with fixed time and memory costs.

\bibliography{__Papers-mix_sample}

\end{document}